\documentclass[journal]{IEEEtran}

\ifCLASSINFOpdf
\else
   \usepackage[dvips]{graphicx}
\fi
\usepackage{url}
\usepackage{graphicx}
\usepackage{cite}
\usepackage{amsmath,amssymb,amsfonts}
\usepackage{bm}
\usepackage{textcomp}
\usepackage{subfigure}
\usepackage{caption}
\usepackage{color}
\usepackage{xcolor}
\usepackage{booktabs}
\usepackage{arydshln}
\usepackage{bbding}
\usepackage{multirow}
\usepackage{diagbox}
\usepackage{ntheorem}
\usepackage{makecell}
\usepackage{colortbl}
\usepackage{pifont}
\usepackage{tablefootnote}
\usepackage[ruled,vlined]{algorithm2e}

\usepackage[pagebackref=true,breaklinks=true,colorlinks,bookmarks=false]{hyperref}

\usepackage[font={bf},textfont=md]{caption}

\usepackage[switch]{lineno} 
\usepackage{siunitx}

\usepackage{array} 
\usepackage{colortbl}%
\newcommand{\myrowcolour}{\rowcolor[gray]{0.925}}
  
\newcommand{\figref}[1]{Fig.~\ref{#1}}
\newcommand{\equref}[1]{Eq.~\eqref{#1}}

\newcommand{\tabref}[1]{Table.~\ref{#1}}

\def\note#1{{\color{black}{{#1}}}}
\def\bluetext#1{{\textcolor{black}{{#1}}}}

\usepackage{xspace}
\makeatletter
\DeclareRobustCommand\onedot{\futurelet\@let@token\@onedot}
\def\@onedot{\ifx\@let@token.\else.\null\fi\xspace}

\def\etal{\emph{et al}\onedot}

\makeatother

\definecolor{seagreen}{RGB}{84,255,159}
\definecolor{SpringGreen}{RGB}{0,139,69}

\newcolumntype{C}[1]{>{\PreserveBackslash\centering}p{#1}}
\newcolumntype{R}[1]{>{\PreserveBackslash\raggedleft}p{#1}}
\newcolumntype{L}[1]{>{\PreserveBackslash\raggedright}p{#1}}

\begin{document}

\title{Decomposed Guided Dynamic Filters for Efficient RGB-Guided Depth Completion}

\author{Yufei Wang, Yuxin Mao, Qi Liu, Yuchao Dai, \IEEEmembership{Member, IEEE}
\thanks{Yufei Wang (wangyufei1951@gmail.com), Yuxin Mao, Qi Liu and Yuchao Dai are with School of Electronics and Information, Northwestern Polytechnical University and Shaanxi Key Laboratory of Information Acquisition and Processing, Xi'an, China. Yuchao Dai (daiyuchao@gmail.com) is the corresponding author.}
}

\markboth{Journal of \LaTeX\ Class Files, Vol. 14, No. 8, August 2015}
{Shell \MakeLowercase{\textit{et al.}}: Bare Demo of IEEEtran.cls for IEEE Journals}

\maketitle

\begin{abstract}
RGB-guided depth completion aims at predicting dense depth maps from sparse depth measurements and corresponding RGB images, where how to effectively and efficiently exploit the multi-modal information is a key issue.
Guided dynamic filters, which generate spatially-variant depth-wise separable convolutional filters from RGB features to guide depth features, have been proven to be effective in this task.
However, the dynamically generated filters require massive model parameters, computational costs and memory footprints when the number of feature channels is large.
In this paper, we propose to decompose the guided dynamic filters into a spatially-shared component multiplied by content-adaptive adaptors at each spatial location.
Based on the proposed idea, we introduce two decomposition schemes $\mathcal{A}$ and $\mathcal{B}$, which decompose the filters by splitting the filter structure and using spatial-wise attention, respectively.
The decomposed filters not only maintain the favorable properties of guided dynamic filters as being content-dependent and spatially-variant, but also reduce model parameters and hardware costs, as the learned adaptors are decoupled with the number of feature channels.
Extensive experimental results demonstrate that the methods using our schemes outperform state-of-the-art methods on the KITTI dataset, and rank 1st and 2nd on the KITTI benchmark at the time of submission.
Meanwhile, they also achieve comparable performance on the NYUv2 dataset.
In addition, our proposed methods are general and could be employed as plug-and-play feature fusion blocks in other multi-modal fusion tasks such as RGB-D salient object detection.
\end{abstract}

\begin{IEEEkeywords}
depth completion, range sensing, guided dynamic filter, multi-modal, feature fusion
\end{IEEEkeywords}

\IEEEpeerreviewmaketitle

\section{Introduction} \label{sec::introduction}

\IEEEPARstart{D}{ense} and accurate depth is essential for various applications, such as obstacle avoidance~\cite{Mur-ArtalT17}, virtual reality~\cite{NewcombeIHMKDKSHF11}, and autonomous driving~\cite{liu2019edge}.
However, current depth sensors are unable to satisfy the requirement for both indoor and outdoor scenes. 
For example, the RGB-D cameras cannot handle transparent and weakly textured areas, and the depth acquired by LiDAR is too sparse to be applied directly.
Therefore, depth completion has been widely studied, which can predict accurate and dense depth maps from available sparse depth measurements.
Since RGB images contain rich structure and semantic cues that are critical for filling unknown depths, using RGB images to guide depth completion (RGB-guided) has become a common paradigm~\cite{van2019sparse, qiu2019deeplidar, ma2018sparse, MaCK19, tang2019learning, HuWLNFG21, Zuo2018edge, Zuo2020depth}.
However, the dense RGB images and sparse depth measurements belong to different modalities~\cite{Zhang2019color, Liu2017depth}. 
How to effectively and efficiently utilize the multi-modal information is a key issue for the RGB-guided depth completion methods.

\begin{figure}
\centering
\begin{minipage}[b]{.45\textwidth}
\centerline{\includegraphics[width=\linewidth]{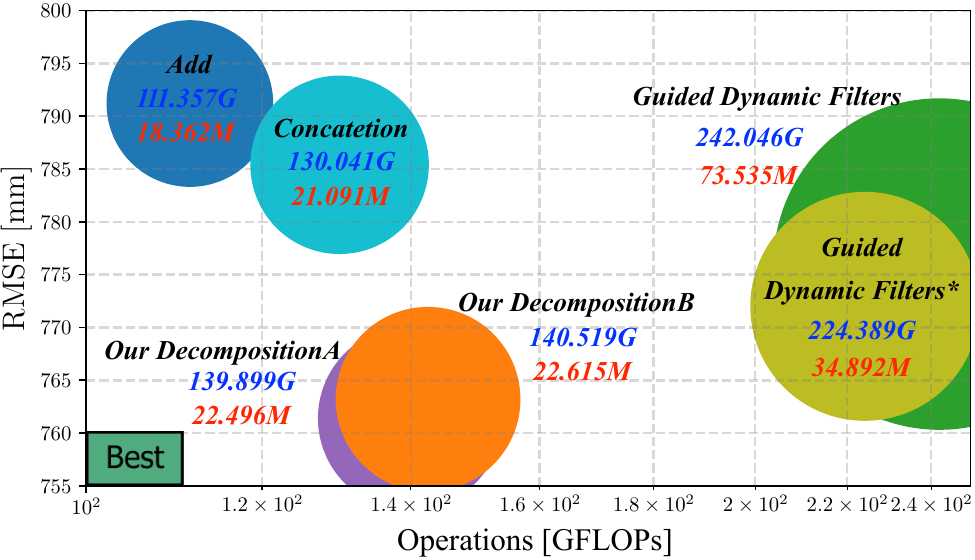}}
\vspace{-1mm}
\caption*{(a) The performance in terms of RMSE versus the computational complexity (denoted in \textcolor{blue}{blue}). The bubble size represents the number of parameters (denoted in \textcolor{red}{red}).}\label{fig:model}
\end{minipage}\qquad 
\vspace{-1mm}
\begin{minipage}[b]{.45\textwidth}
\centerline{\includegraphics[width=\linewidth]{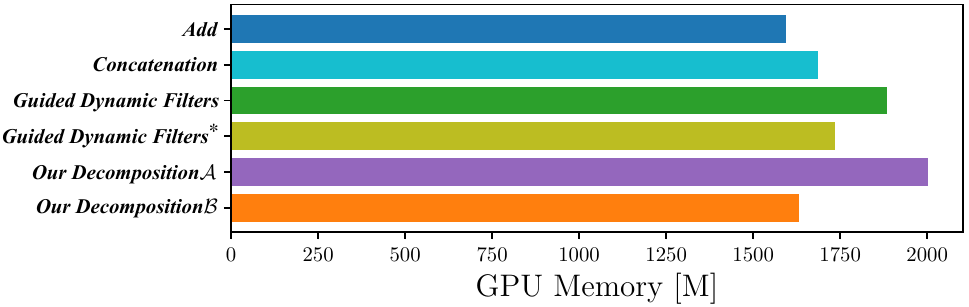}}
\caption*{(b) The memory footprints of different methods.}\label{fig:memory}
\end{minipage}
\caption{{Analysis of GuideNet~\cite{tang2019learning} using different feature fusion methods on the KITTI test dataset}, the methods using our decomposition schemes $\mathcal{A}$ and $\mathcal{B}$ achieve superior performance with smaller model parameters and hardware costs.}
\label{fig:analysis}
\vspace{-5mm}
\end{figure}

\begin{figure*}
\centering
\begin{minipage}[b]{.45\textwidth}
\centerline{\includegraphics[width=\linewidth]{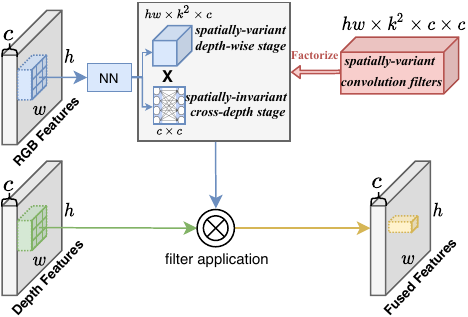}}
\vspace{-1mm}
\caption*{\note{(a) Guided Dynamic Filters}}\label{guidenet}
\end{minipage}\qquad
\vspace{1mm}
\begin{minipage}[b]{.45\textwidth}
\centerline{\includegraphics[width=\linewidth]{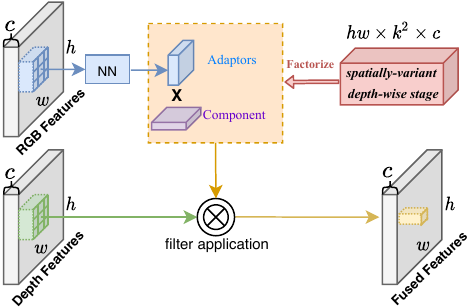}}
\vspace{-1mm}
\caption*{\note{(b) Our Decomposed Guided Dynamic Filters}}\label{decomposed}
\end{minipage}
\vspace{-1mm}
\caption{\note{\textbf{Comparison between guided dynamic filters~\cite{tang2019learning} and our proposed decomposition schemes}. We remove the cross-depth stage of guided dynamic filters and further factorize the depth-wise stage into content-adaptive adaptors and a spatially-shared component. The adaptors are decoupled with the feature channels to make our method more efficient.}}
\label{fig:overview}
\vspace{-3mm}
\end{figure*}

Although existing deep learning-based RGB-guided depth completion methods~\cite{jaritz2018sparse, ma2018sparse, MaCK19,van2019sparse,qiu2019deeplidar} have achieved considerable success by employing diverse network structures, most of them adopt the concatenation or addition operation to fuse the features from the sparse depth and RGB images, which fail to fully utilize the ability of RGB images as guidance~\cite{tang2019learning}.
Inspired by the guided image filtering~\cite{he2012guided} and dynamic filters~\cite{jia2016dynamic}, GuideNet~\cite{tang2019learning} proposes to dynamically generate content-adaptive convolution filters at each spatial location based on the RGB features, and then applies them to guide corresponding depth features, we refer to it as \emph{Guided Dynamic Filters} in this paper.
This approach effectively integrates the RGB and depth information and facilitates training by preventing the gradient from closing to zero~\cite{wu2018dynamic}.
However, generating pixel-wise filters requires prohibitive model parameters, computational costs and memory footprints, which can only be employed in either tiny networks or one layer of networks.
Although this problem can be alleviated by convolution factorization~\cite{howard2017mobilenets} that factorizes the spatially-variant convolution operation into two stages, a spatially-variant depth-wise stage and a spatially-invariant cross-depth stage, the model parameters and hardware costs required by the generated filters still significantly rise with the increase of the number of feature channels.
It will cause an inevitable problem as feature maps in modern networks have hundreds or even thousands of channels~\cite{he_ResNet_CVPR_2016}.

In this paper, we propose Decomposed Guided Dynamic Filters (DGDF) to effectively and efficiently exploit the multi-modal information.
The design of our methods is inspired by the following observations.
Firstly, the spatially-variant depth-wise stage of guided dynamic filters predicts a complete depth-wise convolution filter at each spatial location, which is an obvious over-parameter expression due to the massive spatial redundancy in the images~\cite{subramanya2001image}.
Secondly, the spatially-invariant cross-depth stage of guided dynamic filters requires massive parameters, but its effect is very limited. As shown in \figref{fig:analysis}, guided dynamic filters without this stage, denoted by Guided Dynamic Filters$^{*}$, obtain comparable performance to the original method with smaller model parameters and computational costs.
Therefore, we propose to remove the spatially-invariant cross-depth stage of guided dynamic filters and further decompose the spatially-variant depth-wise stage into a combination of content-adaptive adaptors and a spatially-shared component.
Specifically, as shown in \figref{fig:overview}, we generate the adaptors from the guidance RGB features across the spatial position.
The dimensions of adaptors are decoupled with the number of feature channels and significantly lower than the standard depth-wise filters.
The component is randomly initialized and learned by gradient descent.
Our decomposed guided dynamic filters not only maintain the favorable properties of guided dynamic filters as being spatially-variant and content-adaptive, but also significantly reduce model parameters and hardware costs, which are more friendly to mobile devices.

The key issue of our decomposed guided dynamic filters is how to model the adaptors and the component.
To address this issue, we first propose the decomposition scheme $\mathcal{A}$, which employs content-adaptive bases as the adaptors and expansion coefficients as the component.
The decomposition scheme $\mathcal{A}$ can be easily implemented by two convolution layers.
Meanwhile, it substantially reduces model parameters and computational costs while achieving satisfactory performance.
However, its intermediate feature maps between two convolution layers cost extra memory footprints.
Therefore, we further propose an attention-style decomposition scheme $\mathcal{B}$ to address this problem, which utilizes the standard depth-wise convolution filters as the component and employs the spatial-wise attention as the adopters.
Comprehensive experiments on the KITTI depth completion dataset and NYUv2 dataset verify our methods.
In addition, we conduct extended experiments on the RGB-D salient object detection (SOD) task to demonstrate that our proposed methods are also effective in other multi-modal fusion tasks as plug-and-play feature fusion blocks.

Our contributions can be summarized as follows:
\begin{itemize}
\setlength{\itemsep}{0pt}
\setlength{\parsep}{0pt}
\setlength{\parskip}{0pt}
\item We propose to decompose the guided dynamic filters into a combination of content-adaptive adaptors and a spatially-shared component, which effectively and efficiently exploit the multi-modal information.
\item Two decomposition schemes are proposed to achieve satisfactory accuracy with a significant reduction of model parameters, computational costs, and memory footprints.
\item The methods using our schemes outperform state-of-the-art methods on the KITTI dataset, and rank 1st and 2nd on the KITTI benchmark at the time of submission.
They also achieve comparable performance on the NYUv2 dataset.
Furthermore, our proposed methods are also effective in other multi-modal fusion tasks.
\end{itemize}

\section{Related Work}\label{sec::relatedwork}

\subsection{Depth completion}
Depth completion predicts dense depth maps from sparse depth maps, with optional corresponding RGB images.
RGB-guided depth completion methods usually obtain better performance since they can take advantage of the rich texture and semantic information of the RGB images.
Gansbeke~\etal~\cite{van2019sparse} propose a two-branch network based on RGB images guidance and uncertainty, which achieves precise depth predictions.
Qiu~\etal~\cite{qiu2019deeplidar} consider the low correlation between the RGB images and the depth maps. They propose a method that consists of the surface normal guidance branch and the RGB guidance branch. The proposed method first predicts the surface normal from RGB images and combines the results of two branches by learned confidence maps to obtain final dense depth maps.
Tang~\etal~\cite{tang2019learning} propose guided dynamic filters to effectively fuse the features from RGB images and sparse depth.
Due to the position displacement between the LiDAR and the camera, projecting LiDAR point clouds to the image plane will inevitably cause some foreground and background points to overlap~\cite{zhu2021robust}.
Therefore, some geometry-aware methods~\cite{chen2019learning,zhao2021adaptive,du2022depth} are proposed to obtain better spatial structure information.
In addition, to address the problem that the dense depth maps predicted by end-to-end networks are blurred at the boundaries of objects, a series of spatial propagation networks~\cite{cheng2019cspn,xu2020deformable,park2020non,HuWLNFG21,lin2022dynamic} are proposed to improve the results.

\subsection{Dynamic filters networks}
Dynamic filters networks, which can adjust their structures or weights to different inputs, have been proven to be effective in several tasks~\cite{ma2022generative, jin2022lagconv}. 
Dai~\etal~\cite{dai2017deformable} propose a deformable convolution to adjust receptive fields according to the learned offsets.
Jia~\etal~\cite{jia2016dynamic} introduce dynamic filter networks that predict the filter values by a separate network.
Recent works such as CondConv~\cite{yang2019condconv}, DynamicConv~\cite{chen2020dynamic}, and WeightNet~\cite{ma2020weightnet} generate the dynamic filters by combining several fixed filters. 
The generated filters are based on the input features and are shared spatially.
Since not all spatial locations contribute equally to the final predictions, DRNet~\cite{chen2021dynamic} first predicts several candidate filters according to the input features, and then dynamically selects the most appropriate filter for different spatial locations.
However, spatial or regional shared filters usually lead to sub-optimal results for pixel-wise prediction tasks, as the optimal gradient direction at different pixels may be the same~\cite{su2019pixel}.
Several works~\cite{jia2016dynamic, wang2020solov2} propose to predict a complete convolution filter at each spatial location, but they are restricted by prohibitive model parameters and hardware costs.
To address this issue, many adaptive convolution filters~\cite{su2019pixel,zhou2021decoupled, ZeACDA21} are proposed.
However, these filters are usually used to improve the convolution operation, and their prospects in the multi-modal domain are not fully exploited.

\section{Preliminary} \label{sec:fundamentals}

\subsection{Standard convolution}
The standard convolution operation consists of two steps, namely, neighborhood sampling and aggregation.
\note{Given the input and output feature representations $\mathbf{X},\mathbf{Y} \in \mathbb{R}^{h \times w \times c}$, where $h$ and $w$ are the height and width of the feature map, and $c$ indicates the number of feature channels.
The output feature $\mathbf{Y}_{\mathbf{p}_{i}, \lambda}$ at a spatial position $\mathbf{p}_{i} \in \mathbb{R}^{2}$, and the channel $\lambda \in [1,c]$ can be written as a linear combination of the input features around the location $\mathbf{p}_{i}$}:
\note{
\begin{align}
\footnotesize{\mathbf{Y}_{\mathbf{p}_{i}, \lambda}  = \sum_{\lambda^{\prime} = 1}^{c} \sum_{\mathbf{d}}\!\mathbf{W}_{\mathbf{d} \lambda^{\prime} \lambda} \mathbf{X}_{\mathbf{p}_{i}+\mathbf{d}, \lambda^{\prime}} + \mathbf{b}(\lambda),}
\end{align}
}
\noindent \note{where $\mathbf{d}$ is the deviation in a $k \times k$ sampling grid centered at the position $\mathbf{p}_{i}$ and $\lambda^{\prime}$ is the channel index of the input feature $\mathbf{X}$.}
The parameters of the standard convolution, the filter $\mathbf{W}_{\mathbf{d} \lambda^{\prime} \lambda} \in \mathbb{R}^{k^{2} \times c \times c}$ and the bias $\mathbf{b} \in \mathbb{R}^{c}$, are shared across all spatial locations and different inputs.

\subsection{Dynamic convolution}
Compared with the standard convolution, the dynamic convolution can adaptively adjust filter values at each spatial location $\mathbf{p}_{i}$ according to the input feature $\mathbf{X}$.
Specifically, the dynamic convolution employs an extra network to generate the spatially-variant convolution filter $\mathbf{W}_{\mathbf{p}_{i}\mathbf{d}\lambda^{\prime}\lambda} \in \mathbb{R}^{hw \times k^{2} \times c \times c}$ based on the input feature $\mathbf{X}$, while the standard convolution utilizes fixed parameters at different spatial locations.
Then, the dynamic convolution applies the generated pixel-wise filter back to the input feature, the operation is expressed as:
\note{
 \begin{align}
  &\mathbf{W}_{\mathbf{p}_{i} \mathbf{d} \lambda^{\prime} \lambda} = \mathcal{F}(\mathbf{X}, \Theta), \\
 &\mathbf{Y}_{\mathbf{p}_{i}, \lambda} = 	\sum_{\lambda^{\prime} = 1}^{c} \sum_\mathbf{d} \mathbf{W}_{\mathbf{p}_{i} \mathbf{d} \lambda^{\prime} \lambda} \mathbf{X}_{\mathbf{p}_{i}+\mathbf{d}, \lambda^{\prime}}, 
 \end{align}
}
where $\mathcal{F}$ is the filter generation network parameterized by $\Theta$.
We ignore the bias for convenience of description.
Compared to the standard convolution, the filter $\mathbf{W}_{\mathbf{p}_{i}\mathbf{d}\lambda^{\prime}\lambda}$ is related to the inputs and variants at different locations.

\section{Method}\label{sec::method}
\begin{figure*}[!t]
\centering
\begin{minipage}[b]{.45\textwidth}
\centerline{\includegraphics[width=\linewidth]{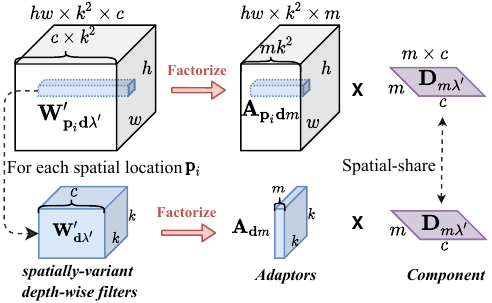}}
\caption*{\note{(a) Our decomposition scheme $\mathcal{A}$.}}\label{fig:decomposedA}
\end{minipage}\qquad ~~
\begin{minipage}[b]{.45\textwidth}
\centerline{\includegraphics[width=\linewidth]{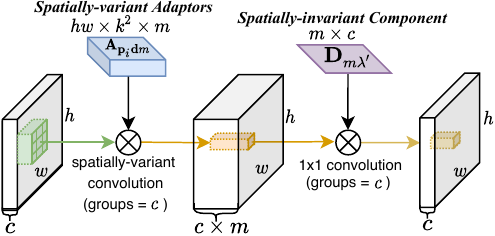}}
\caption*{\note{(b) The two-layer implementation of our scheme $\mathcal{A}$.}}\label{fig:decomposedA-imp}
\end{minipage}
\vspace{-1mm}
\caption{\note{Our decomposition scheme $\mathcal{A}$ decomposes the spatially-variant depth-wise filters $\mathbf{W}_{\mathbf{p}_{i} \mathbf{d} \lambda^{\prime}}^{\prime} \in \mathbb{R}^{hw \times k^{2} \times c}$ into the adaptors $\mathbf{A}_{\mathbf{p}_{i} \mathbf{d} m} \in \mathbb{R}^{hw \times k^{2} \times m}$ and the component $\mathbf{D}_{m\lambda^{\prime}} \in \mathbb{R}^{m \times c}$, and it can be simply implemented by a two-layer convolution. The adaptors are generated from the guidance RGB features, and the component is randomly initialized and learned.}}
\vspace{-3mm}
\label{fig:A}
\end{figure*}

The core of dynamic convolution is to dynamically adjust the filter values for different inputs. 
Notably, the dynamic convolution can be extended to multi-modal tasks, which generates convolution filters from the guidance features, and applies the generated filters to the target features.
For example, in the RGB-guided depth completion task, guided dynamic filters~\cite{tang2019learning} generate spatially-variant and content-adaptive filters from the RGB features, and then employ the generated filters to guide corresponding depth features.
The advantages of guided dynamic filters are two-fold.
Firstly, the filters are spatially-variant, which can not only handle irregular depth features but also prevent the case where the average gradient over all pixels from the next layer is zero~\cite{wu2018dynamic}.
Secondly, the filters are content-adaptive, which can transfer structural details from the RGB images to the depth maps~\cite{he2012guided}, thus fully exploiting the ability of RGB images as guidance.
However, the generated pixel-wise filters require massive model parameters, prohibitive computational costs and memory footprints.

\note{
To alleviate this problem, guided dynamic filters factorize the spatially-variant convolution operation into two stages, the spatially-variant depth-wise stage and spatially-invariant cross-depth stage.
Both stages can be implemented by the convolution operation.
As shown in \equref{equ:gdf}, the convolution filters used in these two stages, denoted as $\mathbf{W}_{\mathbf{p}_{i} \mathbf{d} \lambda^{\prime}}^{\prime}$ and $\mathbf{W}_{\lambda^{\prime} \lambda}^{\prime\prime}$, are dynamically generated based on the guidance RGB features $\mathbf{G} \in  \mathbb{R}^{h \times w \times c}$.
The filter generation network for $\mathbf{W}_{\mathbf{p}_{i} \mathbf{d} \lambda^{\prime}}^{\prime}$, parameterized by $\Theta_1$, consists of two convolution layers, while the filter generation network for $\mathbf{W}_{\lambda^{\prime} \lambda}^{\prime\prime}$, parameterized by $\Theta_2$, first employs the global pooling operation, and then uses two fully connected layers.}
\note{
\begin{equation}
\begin{aligned}
&\mathbf{W}_{\mathbf{p}_{i} \mathbf{d} \lambda^{\prime}}^{\prime} \in \mathbb{R}^{hw \times k^{2} \times c} = \mathcal{F}(\mathbf{G}, \Theta_1), \\
&\mathbf{W}_{\lambda^{\prime} \lambda}^{\prime\prime} \in \mathbb{R}^{c \times c} = \mathcal{F}(\mathbf{G}, \Theta_2).
\end{aligned}
\label{equ:gdf}
\end{equation}
}

\note{Then, guided dynamic filters perform the spatially-variant depth-wise stage and spatially-invariant cross-depth stage through two consecutive convolution layers as:}
\note{
\begin{align}
        &\hat{\mathbf{Y}}_{\mathbf{p}_{i}, \lambda^{\prime}} = \sum_{\mathbf{d}} \mathbf{W}_{\mathbf{p}_{i} \mathbf{d}\lambda^{\prime}}^{\prime} \mathbf{X}_{\mathbf{p}_{i}+\mathbf{d}, \lambda^{\prime}},\\
	&\mathbf{Y}_{\mathbf{p}_{i}, \lambda} = 	\sum_{\lambda^{\prime} = 1}^{c} \mathbf{W}_{\lambda^{\prime} \lambda}^{\prime\prime} \hat{\mathbf{Y}}_{\mathbf{p}_{i},\lambda}.
\end{align}
}

\note{
The ``spatially-variant'' and ``spatially-invariant'' characteristics of these two stages are reflected in whether the filters of two convolution layers vary spatially.
Specifically, the convolution filters $\mathbf{W}_{\mathbf{p}_{i} \mathbf{d} \lambda^{\prime}}^{\prime}$ of the spatially-variant depth-wise stage are variant across different spatial locations $\mathbf{p}_{i}$, while the convolution filters $\mathbf{W}_{\lambda^{\prime} \lambda}^{\prime\prime}$ of the spatially-invariant cross-depth stage are invariant across different locations.}

Although the convolution factorization operation reduces model parameters and hardware costs, the generated filters $\mathbf{W}_{\mathbf{p}_{i} \mathbf{d} \lambda^{\prime}}^{\prime} \in \mathbb{R}^{hw \times k^{2} \times c}$ and $\mathbf{W}_{\lambda^{\prime} \lambda}^{\prime\prime} \in \mathbb{R}^{c \times c}$ are linearly and squarely related to the number of feature channels $c$.
When the feature dimension increases, the computational costs and memory footprints will increase dramatically.
We observe that the spatially-invariant cross-depth stage using two fully connected layers requires massive model parameters, but its effect is similar to the standard $1\times1$ convolution kernel.
As described in Table~\ref{tab:ablation}, guided dynamic filters without this stage perform better than the original method.
Moreover, since the images contain significant spatial redundancy, generating a complete depth-wise filter at each spatial location also suffers from this redundancy, resulting in unnecessary waste of resources.
To address this issue, we propose to remove the spatially-invariant cross-depth stage from the guided dynamic filters and decouple the spatially-variant depth-wise stage with the number of feature channels. We name it Decomposed Guided Dynamic Filters (DGDF).

In this paper, the spatially-variant depth-wise filters are decomposed into a combination of content-adaptive adaptors and a spatially-shared component.
Specifically, as shown in \equref{equ:adaptors} and \equref{equ:component}, the spatially-variant adaptor $\mathbf{A}$ is dynamically generated by a light-weight network from the guidance RGB features, and the spatially-invariant component $\mathbf{D}$ is randomly initialized and learned by gradient descent. 
\begin{align}
 &\mathbf{A} =  \mathcal{F}(\mathbf{G}, \Theta), \label{equ:adaptors} \\
 & \mathbf{D}~ \text{is randomly initialized and learned}.
\label{equ:component}
\end{align}
The decomposed guided dynamic filters maintain the favorable properties of guided dynamic filters with smaller model parameters, computational costs, and memory footprints, as the spatially-variant and content-adaptive filters can be reconstructed by multiplying the spatially-shared component with the content-adaptive adaptors at each spatial location.

\begin{table*}[!t]
\centering
\setlength{\tabcolsep}{1.4 mm}
\caption{{Comparison of model parameters, computational complexity, and memory footprints among different guided dynamic filters, where $N \gg C \gg K^{2} > M$ generally.}}
\label{tab:analy}
\resizebox{1.9\columnwidth}{!}{%
\begin{tabular}{@{}llcccc@{}}
\toprule  
\multicolumn{2}{c}{}                   & Naive Guided Dynamic Filters                     & Guided Dynamic Filters~\cite{tang2019learning}               & Our decomposition scheme $\mathcal{A}$                   & Our decomposition scheme $\mathcal{B}$           \\ \midrule
\multirow{3}{*}{Generation}  & Params. & $C^{3}K^{2}$               & $C^{2}K^{2} + \sigma C^{2} + \sigma C^{3}$ & $CK^{2}M$                                  & $CK^{2}$                   \\
                             & Comput.  & $2NC^{3}K^{2}$             & $2NC^{2}K^{2} + 2\sigma(C^{2} + C^{3})$    & $2NCK^{2}M$                             & $2NCK^{2}$                  \\
                             & Memo.   & $NC^{2}K^{2}$              & $NCK^{2} + \sigma C + C^{2}$                          & $NK^{2}M$                                & $NK^{2}$                  \\ \midrule
\multirow{3}{*}{Application} & Params. & -                          & -                                          & $CM$                               & $CK^{2}$               \\
                             & Comput.  & $2NC^{2}K^{2}$             & $2NCK^{2} + 2NC^{2}$                         & $2NCK^{2}M + 2NCM$            & $3NCK^{2}$             \\
                             & Memo.   & -                          & -                                          & $NCM$                                             & -                          \\ \midrule
\multirow{3}{*}{Summary}     & Params. & $C^{3}K^{2}$               & $C^{2}K^{2} + \sigma C^{2} + \sigma C^{3}$ & $CK^{2}M + CM$             & $2CK^{2}$      \\
                             & Comput.  & $\mathcal{O}(NC^{3}K^{2})$ & $\mathcal{O}(NC^{2}K^{2})$         & $\mathcal{O}(NCK^{2}M)$    & $\mathcal{O}(NCK^{2})$ \\
                             & Memo.   & $\mathcal{O}(NC^{2}K^{2})$ & $\mathcal{O}(NCK^{2})$             & $\mathcal{O}(NCM)$               & $\mathcal{O}(NK^{2})$      \\ \bottomrule
\end{tabular}
}
\end{table*}

\subsection{Decomposition scheme $\mathcal{A}$}
Following the proposed idea, we first propose the decomposition scheme $\mathcal{A}$, which decomposes the spatially-variant depth-wise filters $\mathbf{W}_{\mathbf{p}_{i} \mathbf{d}\lambda^{\prime}} \in \mathbb{R}^{hw \times k^{2} \times c}$ into the product of content-adaptive bases with the coefficients.
Specifically, at each spatial location, we employ an auxiliary network to generate $m$ content-adaptive bases from the guidance RGB features.
The coefficients used in the combination are randomly initialized and then updated by learning.
It is worth noting that these coefficients are shared in the spatial domain.
In our decomposition scheme $\mathcal{A}$, the dynamically generated bases are used as the adaptors $\mathbf{A}_{\mathbf{p}_{i} \mathbf{d} m} \in \mathbb{R}^{hw \times k^{2} \times m}$, and the spatially-shared component is used as the component $\mathbf{D}_{m\lambda^{\prime}} \in \mathbb{R}^{m \times c}$.
As shown in \figref{fig:A}~(a), the guided dynamic filters at each spatial location $\mathbf{W}_{\mathbf{p}_{i} \mathbf{d}\lambda^{\prime}}^{\prime}$ can be reconstructed by multiplying the adaptors $\mathbf{A}_{\mathbf{p}_{i} \mathbf{d} m}$ with the component $\mathbf{D}_{m\lambda^{\prime}}$.
Compared with the typical guided dynamic filters $\{\mathbf{W}_{\mathbf{p}_{i} \mathbf{d} \lambda^{\prime}}^{\prime} \in \mathbb{R}^{hw \times k^{2} \times c}, \mathbf{W}_{\lambda^{\prime} \lambda}^{\prime\prime} \in \mathbb{R}^{c \times c}\}$, our decomposed guided dynamic filters $\{\mathbf{A}_{\mathbf{p}_{i} \mathbf{d} m} \in \mathbb{R}^{hw \times k^{2} \times m}, \mathbf{D}_{m\lambda^{\prime}} \in \mathbb{R}^{m \times c}\}$ significantly reduce the number of model parameters and their accompanying hardware costs, as $m \ll c$ and $\mathbf{D}_{m\lambda^{\prime}}$ is not generated by the network.
In addition, Fig.~\ref{fig:A}~(b) demonstrates that our proposed decomposition scheme $\mathcal A$ can be easily implemented by a two-layer convolution, where the depth features are first convoluted with $m$ adaptors channel by channel, and then convoluted with the component.

\subsection{Decomposition scheme $\mathcal{B}$}
\begin{figure}[h]
	\centering
	\includegraphics[width=0.45\textwidth]{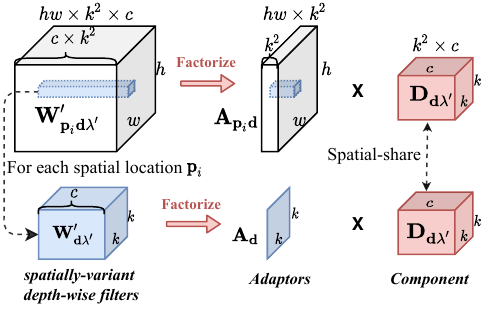}
	\caption{\note{Our decomposition scheme $\mathcal{B}$ decomposes the spatially-variant depth-wise filters $\mathbf{W}_{\mathbf{p}_{i} \mathbf{d} \lambda^{\prime}}^{\prime} \in \mathbb{R}^{hw \times k^{2} \times c}$ by applying spatial-wise attention maps $\mathbf{A}_{\mathbf{p}_{i} \mathbf{d}} \in \mathbb{R}^{hw \times k^{2}}$ to a static depth-wise filter $\mathbf{D}_{\mathbf{d} \lambda^{\prime}} \in \mathbb{R}^{k^{2} \times c}$ at each spatial location.
 }}
	\label{fig:schemeB}
	\vspace{1mm}
\end{figure}

Although the proposed decomposition scheme $\mathcal{A}$ effectively exploits the RGB images and sparse depth information with small model parameters and computational costs, the intermediate feature map in its two-layer implementation has $cm$ channels.
When the number of feature channels $c$ is large, it still costs heavy memory footprints.
To address this issue, we further propose an attention-style decomposition scheme $\mathcal{B}$.
We observe that the spatially-variant depth-wise filters $\mathbf{W}_{\mathbf{p}_{i} \mathbf{d} \lambda^{\prime}}^{\prime} \in \mathbb{R}^{hw \times k^{2} \times c}$ is equivalent to applying a spatially-variant spatial attention map to a standard static depth-wise filter at each spatial location.
As shown in Fig.~\ref{fig:schemeB}, the proposed decomposition scheme $\mathcal{B}$ employs spatial-wise attention maps as the adaptors, namely $\mathbf{A}_{\mathbf{p}_{i} \mathbf{d}} \in \mathbb{R}^{hw \times k^{2}}$, which are dynamically predicted by an extra network based on the guidance RGB features.
A standard static depth-wise filter is used as the component $\mathbf{D}_{\mathbf{d} \lambda^{\prime}} \in \mathbb{R}^{k^{2} \times c}$ of the decomposition scheme $\mathcal{B}$.
Different from the decomposition scheme $\mathcal{A}$, our decomposition scheme $\mathcal{B}$ can be implemented through one-layer convolution with spatial-wise attention.
It does not contain intermediate feature maps, which can further reduce the memory footprints.
\note{It is worth noting that the decomposition scheme $\mathcal{B}$ is actually a special case of the decomposition scheme $\mathcal{A}$. When the decomposition scheme $\mathcal{A}$ uses only one base as the adaptor and expands the coefficients to pixel-wise, the decomposition scheme $\mathcal{A}$ is equal to the decomposition scheme $\mathcal{B}$.}

\begin{figure*}[!t]
	\centering
	\includegraphics[width=0.95\textwidth]{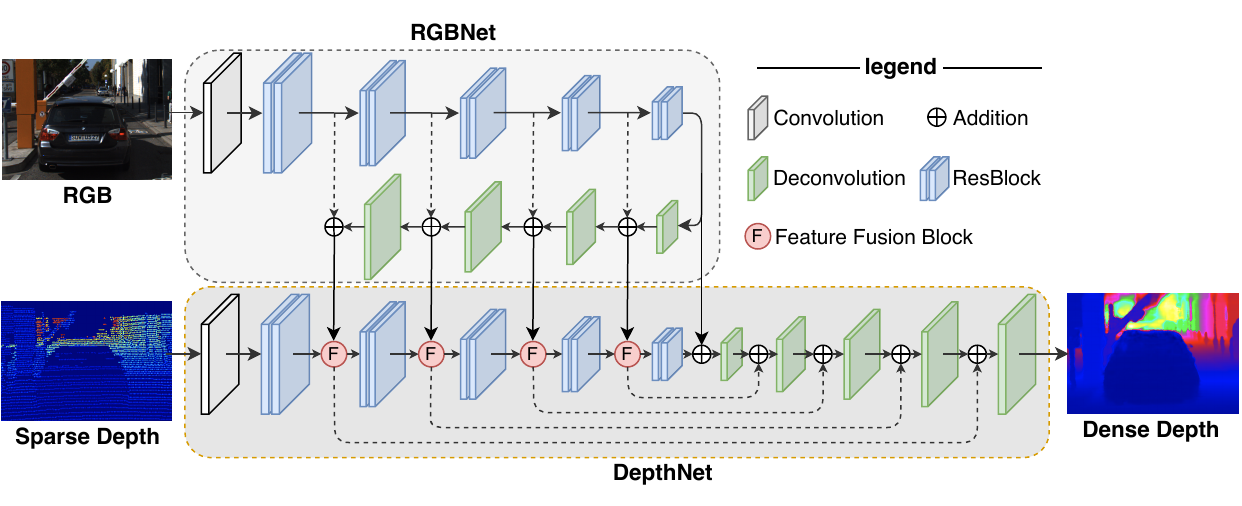}
        \vspace{-4mm}
	\caption{The depth completion network architecture used by our proposed method, which consists of two sub-networks, RGBNet and DepthNet. The multi-scale RGB features and depth features are fused through our proposed methods.}
	\label{fig:method}
	\vspace{-1mm}
\end{figure*}

\subsection{Complexity analysis}

In this subsection, we compare the model parameters, computational complexity, and memory footprints among different feature fusion methods, including guided dynamic filters with and without the convolution factorization operation, and our decomposition schemes $\mathcal{A}$ and $\mathcal{B}$.
We denote guided dynamic filters without the convolution factorization operation as naive guided dynamic filters.
Suppose the shape of all used feature maps is $H \times W \times C$, where $H$ and $W$ are the height and width, and $C$ is the number of feature channels. 
$N=H \times W$ is the number of pixels, $K$ is the size of the guidance filter, $\sigma$ is the squeeze ratio used in the convolution factorization operation, and $M$ is the number of bases in our decomposition scheme $\mathcal{A}$.

All methods involved in the comparison consist of two modules, generating filters or adaptors from the guidance features, and applying the generated or reconstructed filters to the target features.
For simplicity, we assume all the generated networks adopt a convolution layer with a $1 \times 1$ filter size.
For example, the naive guided dynamic filters generate a complete convolution filter $C^{2}K^{2}$ at each spatial location, which requires $C^{3}K^{2}$ model parameters and $2NC^{3}K^{2}$ floating-point operations (FLOPs), and the generated filters cost $NC^{2}K^{2}$ memory. Guided dynamic filters split the convolution operation into two stages, whose filters are of shape $NCK^{2}$ and $C^{2}$.
Among them, the second stage using a squeeze-and-excitation layer~\cite{hu2018squeeze} has $\sigma C^{2}$ parameters for the squeeze layer and $\sigma C^{3}$ parameters for the excitation layer, which takes $2\sigma(C^{2} + C^{3})$ FLOPs.

For the filter application module, naive guided dynamic filters and guided dynamic filters can directly employ the generated filters to the target features.
They do not require additional parameters, while our decomposition schemes need to employ their respective components to reconstruct the spatially-variant filters.
The component parameters contained in our decomposition schemes $\mathcal{A}$ and $\mathcal{B}$ are $CM$ and $CK^{2}$, respectively. 
In this module, naive guided dynamic filters using the standard convolution operation take $2NC^{2}K^{2}$ FLOPs, and guided dynamic filters using the depth-wise separable convolution operation~\cite{howard2017mobilenets} costs $2NCK^{2} + 2NC^{2}$ FLOPs.
Our decomposition scheme $\mathcal{A}$ is implemented by two-layer convolution, which takes $2NCK^{2}M + 2NCM$ FLOPs, and the intermediate feature costs $NCM$ memory.
Our decomposition scheme $\mathcal{B}$ employs the depth-wise convolution with spatially-variant attention, which takes $3NCK^{2}$ FLOPs.

The comparison results shown in Table~\ref{tab:analy} demonstrate that, compared to guided dynamic filters with and without the convolution factorization operation, our proposed decomposed guided dynamic filters significantly reduce the model parameters, computational complexity, and memory footprints.

\subsection{Implementation details}

\noindent\textbf{Network architecture.} 
As shown in Fig.\ref{fig:method}, our depth completion network employs a double encoder-decoder structure, which consists of two sub-networks, RGBNet and DepthNet.
Our proposed decomposed guided dynamic filters are inserted into the network as a plug-and-play feature fusion block.
Specifically, the RGB images are fed into the RGBNet to learn multi-scale RGB features and then utilized to guide the corresponding scale depth features extracted by the DepthNet.

\noindent\textbf{Loss function.} We employ an $L_2$ loss as: 
\begin{align}
\mathcal{L}(d^{pred})=\| (d^{pred} - d^{gt}) \odot \mathbf{1}_{\{d^{gt} > 0\}}  \|^{2},
\label{equ:l2}
\end{align}
where $d^{pred}$ is the predicted dense depth map, $d^{gt}$ is the ground truth depth map. Since the ground truth depth maps are usually semi-dense, we only supervise the available parts, $\mathbf{1}_{\{d^{gt} > 0\}}$ indicates whether there is a value in the ground truth, $\odot$ denotes the element-wise multiplication.

\noindent\textbf{Training details.} Our method is implemented by PyTorch and trained on 2$\times$2080Ti GPUs for 30 epochs. All experiments are conducted by the AdamW optimizer with $\beta_{1} = 0.9$, $\beta_{2} = 0.999$.
The size of the mini-batch is 8 and the initial learning rate is $10^{-3}$. The learning rate is then reduced by $50\%$ every 5 epochs.

\section{Experiments}\label{sec:experiment}

\begin{table*}[!t]
\centering
\renewcommand\arraystretch{1.1}
\caption{\note{Performance comparison of GuideNet~\cite{tang2019learning} using different feature fusion methods on the KITTI validation dataset. 
Guided dynamic filters$^{*}$ represent the guided dynamic filters without the spatially-invariant cross-depth stage.  
The decomposition scheme $\mathcal{B}$ using the channel attention map is denoted as ``our decomposition $\mathcal{B}$\_channel''.
``$5 \times 5$'' represents the kernel size of the guided dynamic filter, and the default kernel size of other guided dynamic filters is $3 \times 3$.
 }}
 \vspace{-1mm}
\resizebox{\textwidth}{!}{%
\begin{tabular}{cccccccc}
\toprule
    Feature Fusion Methods    & RMSE[mm]   & MAE[mm]   & iRMSE[1/km] & iMAE[1/km] & Parameters[Million] & Memory[MB] & Speed[s] \\ \midrule 
Add      & 791.2  & 220.3 & 2.4   & 1.0  & 18.4M  & 1595M  & 0.03s \\
Concatenation   & 785.4  & 221.9 & 2.3   & 1.0  & 21.1M  & 1687M  & 0.03s \\
Guided dynamic filters~\cite{tang2019learning}    & 776.0  & 219.0 & 2.2   & 1.0  & 73.5M  & 1885M  & 0.08s \\ 
Guided dynamic filters*    & 772.0  & 219.0 & 2.3   & 1.0  & 34.9M  & 1735M  & 0.08s \\ \hline
\myrowcolour Our decomposition $\mathcal{A}$   & 761.4  & 218.9 & 2.3   & 1.0  & 22.5M  & 2003M  & 0.05s \\
\myrowcolour Our decomposition $\mathcal{B}$ & 762.2  & 216.0 & 2.2   & 1.0  & 22.6M  & 1631M  & 0.06s \\ \hline
\bluetext{Our decomposition $\mathcal{B}$\_channel} & \bluetext{765.9}  & \bluetext{215.7} & \bluetext{2.2}   & \bluetext{1.0}  & \bluetext{22.8M}  & \bluetext{1791M}  & \bluetext{0.06s} \\ 
\bluetext{Our decomposition $\mathcal{A}$ (5$\times$5)}   & \bluetext{763.0}  & \bluetext{217.2} & \bluetext{2.2}   & \bluetext{1.0}  & \bluetext{22.6M}  & \bluetext{2861M}  & \bluetext{0.05s} \\
\bluetext{Our decomposition $\mathcal{B}$ (5$\times$5)} & \bluetext{763.5}  & \bluetext{215.0} & \bluetext{2.2}   & \bluetext{1.0}  & \bluetext{22.6M}  & \bluetext{1874M}  & \bluetext{0.06s} \\ 
\bottomrule
\end{tabular}
}
\label{tab:ablation}
\vspace{-2mm}
\end{table*}

\subsection{Datasets and Metrics}

\noindent\textbf{KITTI Dataset~\cite{Geiger2012CVPR}.}
The KITTI dataset~\cite{Geiger2012CVPR} is a large real-world autonomous driving dataset, which consists of sparse depth maps obtained by a Velodyne 64-line LiDAR and corresponding RGB images.
The ground truth depth maps are created by aggregating 11 consecutive LiDAR scans into one.
Same as existing methods, we employ 86k images for training and evaluate the performance on 1k selected validation images.
The dataset also provides 1k images without ground truth that need to be tested on the KITTI online benchmark for a fair comparison.
Since the top of the sparse depth map does not have valid LiDAR points, we crop the input images to $256\times1216$ for both training and inference as \cite{van2019sparse,tang2019learning}.

\noindent\textbf{NYUv2 Dataset~\cite{Silberman12}.}
The NYUv2 dataset consists of RGB images and depth maps obtained from 464 different indoor scenes.
The depth maps are acquired by a Microsoft Kinect camera.
Following the same setting of previous depth completion methods~\cite{MaCK19,cheng2019cspn, qiu2019deeplidar}, we train the model with 50K images uniformly sampled from the training set and test it on 654 officially labeled images.
As a pre-processing, the depth images are in-painted by the official toolbox to fill in the missing values.
For both train and test datasets, the original images of size $640\times 480$ are downsampled to half and then center-cropped to $304\times 228$. 

\noindent\textbf{Evaluation metrics.}
Following exiting depth completion methods, we employ the root mean squared error (RMSE[$\mathrm{mm}$]), mean absolute error (MAE[$\mathrm{mm}$]), root mean squared error of the inverse depth (iRMSE[1 / $\mathrm{km}$]), mean absolute error of the inverse depth (iMAE[1 / $\mathrm{km}$]), relative absolute error (REL), and $\delta$ inlier ratios (maximal mean relative error of $\delta_\tau=1.25^{\tau}$ for $\tau \in {1,2,3}$) for quantitative evaluation. 
Eq.~\eqref{equ:metrics} shows their detailed definitions, where $d^{gt}$ denotes the ground truth, $d^{pred}$ denotes the predicted dense depth map, and $\mathcal{V}$ is the set of the available points in the ground truth.

\begin{equation}
\label{equ:metrics}
\begin{aligned}
&\operatorname{{RMSE~}} {[{\mathrm{mm}}]}:~ \sqrt{\frac{1}{|\mathcal{V}|} \sum_{v \in \mathcal{V}}\left|d_{v}^{gt}-d_{v}^{pred}\right|^{2}}, \\
&\operatorname{{MAE~}} {[{\mathrm{mm}}]}:~  \frac{1}{|\mathcal{V}|} \sum_{v \in \mathcal{V}}\left|d_{v}^{g t}-d_{v}^{ {pred }}\right|, \\
&\operatorname{{iRMSE~}} {[\mathrm{1/{km}}]}:~ \sqrt{\frac{1}{|\mathcal{V}|} \sum_{v \in \mathcal{V}}\left|1 / d_{v}^{g t}-1 / d_{v}^{ {pred }}\right|^{2}}, \\
& \operatorname{{iMAE~}} {[\mathrm{1/{km}}]}:~ \frac{1}{|\mathcal{V}|} \sum_{v \in \mathcal{V}}\left|1 / d_{v}^{g t}-1 / d_{v}^{ {pred }}\right|, \\
& \operatorname {{REL~}}:~ \frac{1}{|\mathcal{V}|} \sum_{v \in \mathcal{V}}\left|\left(d_v^{g t}-d_v^{p r e d}\right) / d_v^{g t}\right|, \\
&{\delta_\tau }[\%]:~  \max \left(\frac{d_v^{g t}}{d_v^{p r e d}}, \frac{d_v^{p r e d}}{d_v^{g t}}\right)<\tau.
\end{aligned}
\end{equation}

\begin{table*}[!t]
\centering\
\renewcommand\arraystretch{1.1}
\caption{Ablation studies on KITTI validation dataset. Stochastic depth means the stochastic depth training strategy.}
\vspace{1mm}
\label{tab:ablation_stydies}
\begin{tabular}{@{}cccccccccc@{}}
\toprule
\multirow{2}{*}{Methods} &
  \multicolumn{4}{c}{Multi-scale Fusion Scheme} &
  \multirow{2}{*}{\begin{tabular}[c]{@{}c@{}}~Stochastic  depth~\end{tabular}} &
  \multirow{2}{*}{\begin{tabular}[c]{@{}c@{}}~RMSE~ \\ {[}mm{]}\end{tabular}} &
  \multirow{2}{*}{\begin{tabular}[c]{@{}c@{}}~MAE~ \\ {[}mm{]}\end{tabular}} &
  \multirow{2}{*}{\begin{tabular}[c]{@{}c@{}}~iRMSE~\\ {[}1/km{]}\end{tabular}} &
  \multirow{2}{*}{\begin{tabular}[c]{@{}c@{}}~iMAE~~~~~\\ {[}1/km{]~}\end{tabular}} \\ \cmidrule(lr){2-5}
  & ~~~full-scale   & 1/2-scale    & 1/4-scale    & 1/8-scale    &              &       &       &     &     \\ \midrule
~~~Our \emph{DGDF-A} &              & $\checkmark$ & $\checkmark$ & $\checkmark$ &              & 761.4 & 218.9 & 2.3 & 1.0 \\
~~~Our \emph{DGDF-B} &              & $\checkmark$ & $\checkmark$ & $\checkmark$ &              & 762.2 & 216.0 & 2.2 & 1.0 \\ 
~~~Our \emph{DGDF-A} & $\checkmark$ & $\checkmark$ & $\checkmark$ & $\checkmark$ &              & 754.2 & 208.0 & 2.1 & 0.9 \\
~~~Our \emph{DGDF-B} & $\checkmark$ & $\checkmark$ & $\checkmark$ & $\checkmark$ &              & 752.4 & 210.8 & 2.1 & 0.9 \\ 
~~~Our \emph{DGDF-A} & $\checkmark$ & $\checkmark$ & $\checkmark$ & $\checkmark$ & $\checkmark$ & 746.5 & 206.0 & 2.2 & 0.9 \\
~~~Our \emph{DGDF-B} & $\checkmark$ & $\checkmark$ & $\checkmark$ & $\checkmark$ & $\checkmark$ & 739.6 & 205.2 & 2.1 & 0.9 \\ \bottomrule
\end{tabular}
\vspace{-1mm}
\end{table*}

\begin{figure*}[htbp]
	\centering
	\includegraphics[width=0.9\textwidth]{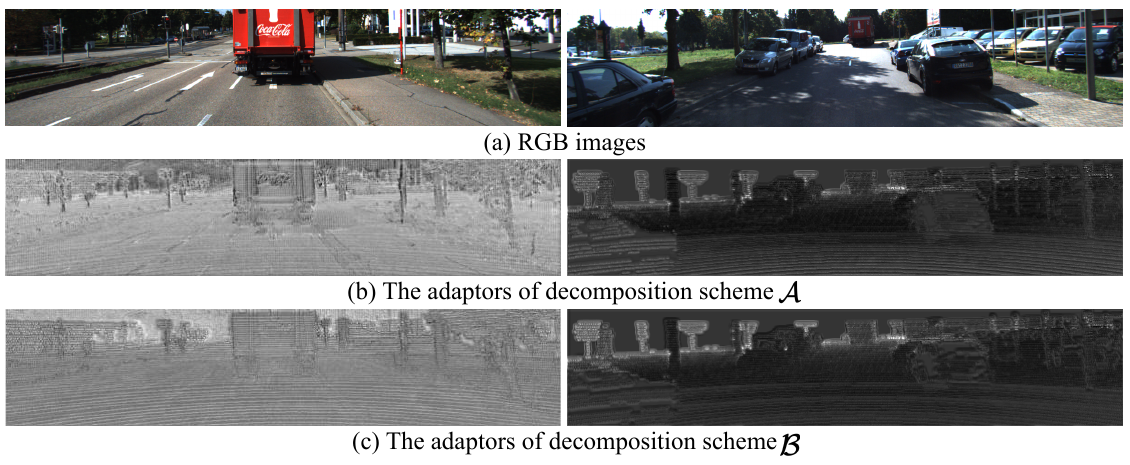}
        \vspace{-1mm}
	\caption{{Visualization of the learned adaptors that are spatially-variant and content-adaptive.}}
	\label{fig:adaptor}
	\vspace{-3mm}
\end{figure*}

\subsection{Ablation studies}
\label{sec:ablation}

To verify the effectiveness of various components in our proposed method, we conduct extensive ablation studies on the KITTI validation dataset.
Specifically, we first compare the performance of the RGB-guided depth completion method using different feature fusion schemes.
Then, we verify the robustness of our methods under various input depth densities.
In addition, we investigate the effectiveness of the full-scale feature fusion and stochastic depth training strategy.
We denote the depth completion methods that use our decomposition schemes $\mathcal{A}$ and $\mathcal{B}$ as \emph{\textbf{DGDF-A}} and \emph{\textbf{DGDF-B}}, respectively.

\noindent\textbf{Comparison of feature fusion methods.}
We use the network architecture of GuideNet~\cite{tang2019learning} as the backbone.
In Table~\ref{tab:ablation}, we compare the method using different feature fusion modules in terms of performance, model parameters, memory footprints, and speed.
For a fair comparison, we only replace the feature fusion block of the network and maintain other network components and settings unchanged.
The experimental results demonstrate that, compared with the element-wise addition (denoted as Add), the channel-wise concatenation (denoted as Concatenation) achieves better results, as Concatenation is generally considered to be more appropriate for fusing homogeneous features.
Guided dynamic filters, consisting of a spatially-variant depth-wise stage and a spatially-invariant cross-depth stage, effectively improve the performance of the method.
However, it suffers from heavy model parameters, computational costs, and memory footprints.
\note{Meanwhile, we observe that guided dynamic filters without the spatially-invariant cross-depth stage (denoted as guided dynamic filters$^{*}$) perform better with smaller model parameters and hardware costs. 
Since the gradient of a spatially-invariant filter is more likely to be close to zero~\cite{tang2019learning, wu2018dynamic}, the spatially-invariant stage may adversely affect the training process, which impairs the performance of the guided dynamic filters.}

In this paper, we propose two decomposed guided dynamic filters $\mathcal{A}$ and $\mathcal{B}$ to efficiently exploit dense RGB images and sparse depth maps.
Our method removes the spatially-invariant cross-depth stage from the guided dynamic filters and decomposes the spatially-variant depth-wise stage into the combination of content-adaptive adaptors and a spatially-shared component.
The results demonstrate that, compared to guided dynamic filters, our decomposition scheme $\mathcal{A}$ improves the RMSE accuracy by 14.6mm, with the model parameters decreasing 51M and the speed decreasing 0.03s respectively.
Meanwhile, the proposed decomposition scheme $\mathcal{B}$ significantly reduces the memory footprints  while maintaining the performance, which uses spatially-variant channel attention maps to a standard static depth-wise filter at each location. 

\note{In addition, we also show the results of the decomposition scheme $\mathcal{B}$ using the spatially-variant channel attention maps, which is denoted as ``our decomposition $\mathcal{B}$\_channel''.
The results illustrate that the decomposition scheme $\mathcal{B}$ using the channel attention map obtains comparable results with the decomposition scheme $\mathcal{B}$ using the spatial attention map, but requires more model parameters and memory footprints.
In this paper, the kernel sizes of all guide dynamic filters are set to $3 \times 3$ by default.
We enlarge the kernel size of our decomposed guided dynamic filters $\mathcal{A}$ and $\mathcal{B}$ to $5 \times 5$ to verify whether enlarging the kernel size of the guide dynamic filters will bring the performance gains.
The results demonstrate that the guided dynamic filters with larger kernel size do not improve the results, but consume more model parameters and hardware costs.
We consider that selecting the kernel size of the guided dynamic filters is a trade-off, the large kernel size increases the receptive field while more likely to introduce irrelevant information.
Therefore, we still set the kernel size of our decomposed guided dynamic filters $\mathcal{A}$ and $\mathcal{B}$ to $3 \times 3$.
}

To verify whether the adaptors of our decomposed guided dynamic filters are spatially-variant and content-adaptive, we visualize them in Fig.~\ref{fig:adaptor}.
Since the adaptors have multiple channels, we add up the values of all channels and scale the sum to 0-1.
We observe that the adaptors predicted by our proposed decomposition schemes $\mathcal{A}$ and $\mathcal{B}$ are spatially different and correlate with the RGB image content, which demonstrates our method maintains the favorable properties of guided dynamic filters.

\begin{figure}[!t]
\vspace{-1mm}
	\centering
	\includegraphics[width=0.45\textwidth]{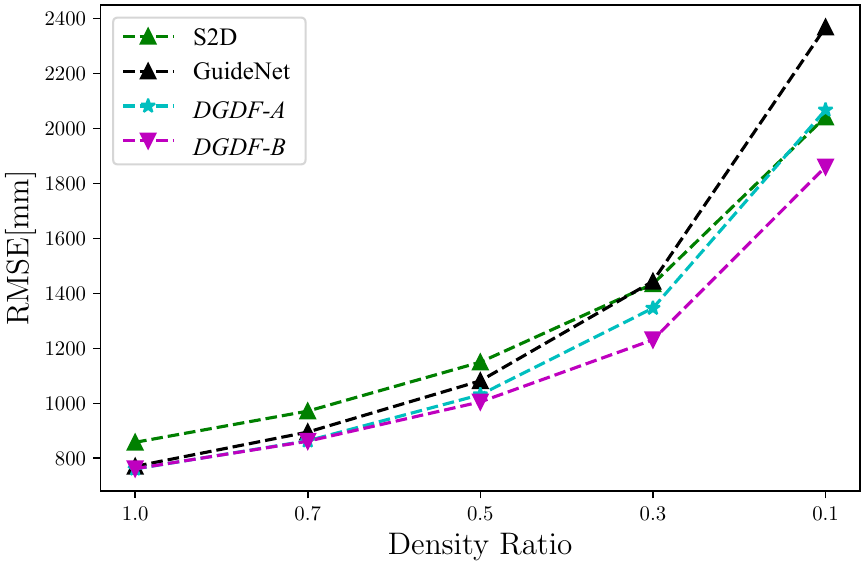}
	\caption{{Performance comparison in term of RMSE[$\mathrm{mm}$] under different levels of input depth density.}}
	\label{fig:sparse}
	\vspace{-3mm}
\end{figure}

\begin{figure*}
	\centering
	\includegraphics[width=1.0\textwidth]{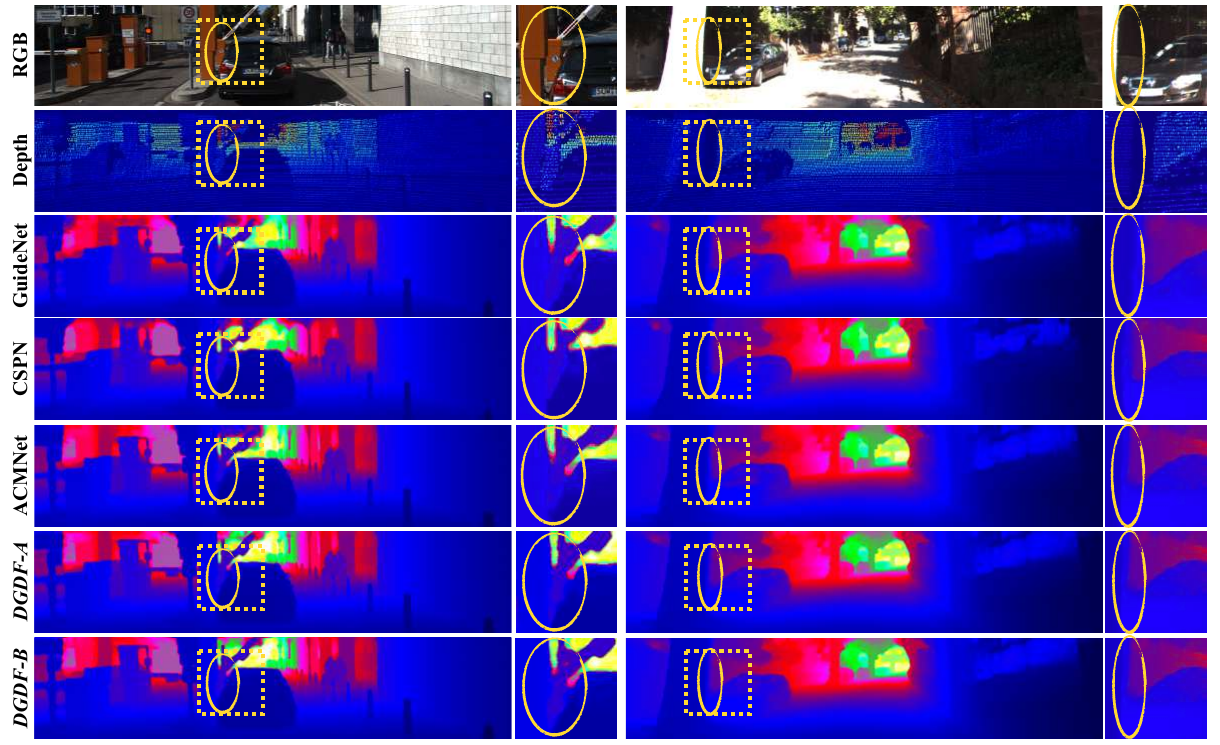}
	\caption{{Qualitative comparison with state-of-the-art methods on the KITTI test dataset. From top to bottom are RGB images, sparse input depth maps, the dense depth maps predicted by GuideNet~\cite{tang2019learning}, CSPN~\cite{cheng2019cspn}, ACMNet~\cite{zhao2021adaptive} and our \emph{DGDF-A} and \emph{DGDF-B}, respectively. We zoom in some representative areas for detailed comparison.}}
	\label{fig:Qualitative}
	\vspace{-3mm}
\end{figure*}

\noindent\textbf{Robust to different input depth densities.}
The sparse depth maps of the KITTI dataset~\cite{Geiger2012CVPR} are obtained by a 64-line LiDAR. 
However, in many practical applications, only 32-line or 16-line LiDAR will be employed due to the cost constraint, which only provides more sparse input depth maps.
Therefore, it is crucial to analyze the performance of the proposed methods on sparse depth maps with different sparsity levels.
By randomly sampling input sparse depth maps, we generate more sparse depth maps according to different ratios of density.
Our method also uses the network of GuideNet~\cite{tang2019learning} as the backbone.
Fig.~\ref{fig:sparse} compares the performance of our approaches with S2D~\cite{MaCK19} and GuideNet~\cite{tang2019learning} under different sparsity levels on the KITTI validation dataset.
The results demonstrate that the performance of all methods drops dramatically with the depth density decreasing.
Our \emph{DGDF-A} performs better than S2D and GuideNet except when the density ratio is 0.1, while our \emph{DGDF-B} outperforms other methods at all density ratios.
Experimental results illustrate that our methods have powerful robustness for depth maps with different levels of sparsity.

\noindent\textbf{Full-scale feature fusion.}
The multi-scale feature fusion scheme has been proven to be effective in GuideNet~\cite{tang2019learning}.
However, guided dynamic filters require massive model parameters and hardware costs, GuideNet only employs the feature fusion method to fuse small-scale RGB and depth feature maps, such as the feature maps of 1/2-scale, 1/4-scale and 1/8-scale.
Since the proposed decomposed guided dynamic filters are more effective and efficient, they can not only fuse small-scale feature maps well, but also be used to fuse full-scale feature maps with limited resources.
Table~\ref{tab:ablation_stydies} demonstrates that the full-scale feature fusion scheme effectively improves the performance of the method, as the full-scale feature map contains more detailed structures.

\noindent\textbf{Stochastic depth training strategy.}
The stochastic depth training strategy~\cite{Huang2016Deep} is proposed to improve the training speed and method performance.
It randomly deactivates some layers in the network as the Dropout, so that the final model is a combination of models with different depth.
We adopt the stochastic depth strategy to train our models as \cite{lin2022dynamic}.
The experiment results in Table~\ref{tab:ablation_stydies} demonstrate that the training strategy effectively improves the performance of the method.

\begin{figure*}[!t]
	\centering
	\includegraphics[width=0.97\textwidth]{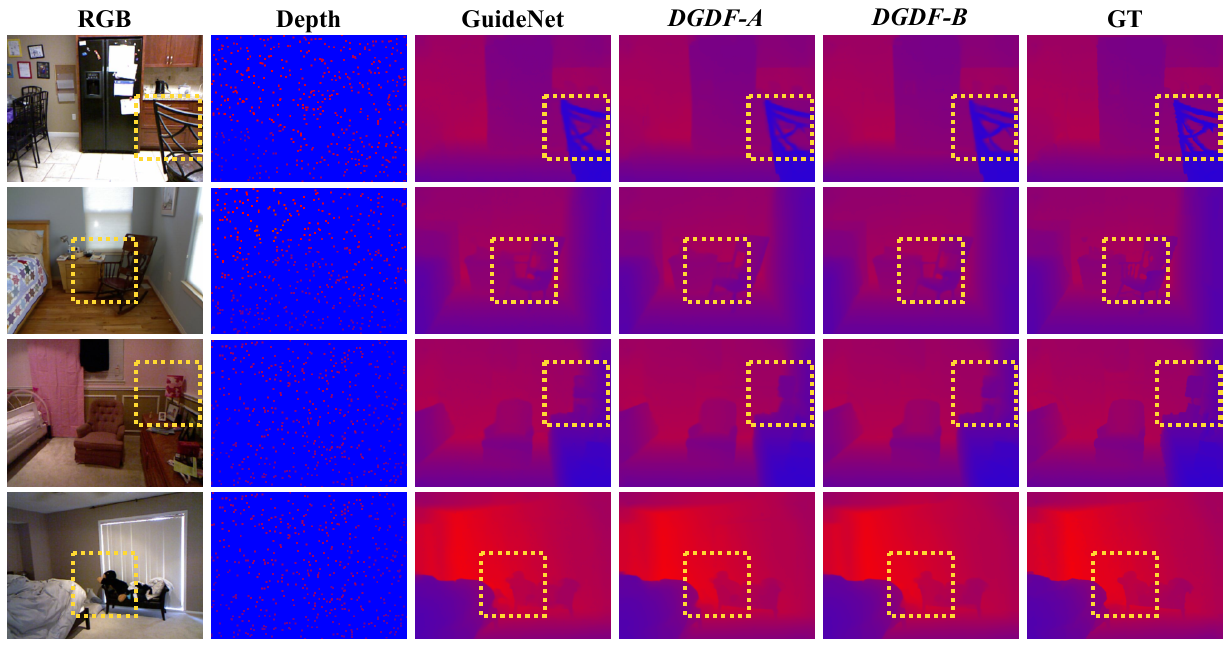}
	\caption{{Qualitative evaluation on the NYUv2 dataset. From left to right are RGB images, sparse input depth maps, the dense depth maps predicted by GuideNet~\cite{tang2019learning}, our \emph{DGDF-A} and \emph{DGDF-B}, and the ground truth.}}
	\label{fig:nyuv2}
	\vspace{-2mm}
\end{figure*}

\begin{table}[!t]
\centering
\renewcommand\arraystretch{1.2}
\caption{Quantitative comparison with state-of-the-art depth completion methods on the KITTI leaderboard. The best and second-best results are highlighted in \textcolor{red}{red} and \textcolor{blue}{blue} colors, respectively.
}
\label{tab:Quantitative}
\begin{tabular}{lcccc}
\hline
\myrowcolour
    		Methods & \begin{tabular}[c]{@{}c@{}}\textbf{RMSE} \\ {[}mm{]}\end{tabular}   &
                \begin{tabular}[c]{@{}c@{}}MAE \\ {[}mm{]}\end{tabular} 
                &\begin{tabular}[c]{@{}c@{}}iRMSE~ \\ {[}1/km{]}\end{tabular} & 
                \begin{tabular}[c]{@{}c@{}}iMAE~ \\ {[}1/km{]}\end{tabular} \\ \hline 
    		CSPN~\cite{cheng2019cspn}    & 1019.64 & 279.46 & 2.93  & 1.15 \\
    		S2D~\cite{MaCK19}     & 814.73  & 249.95 & 2.80  & 1.21 \\ 
                DepthNormal~\cite{xu2019depth}  & 777.05 & 235.17 & 2.42 & 1.13 \\
    		GAENet~\cite{du2022depth}  & 773.90  & 231.29 & 2.29  & 1.08 \\
    		Uncertainty~\cite{van2019sparse} & 772.87 & 215.02 & 2.19 & 0.93 \\
    		DeepLiDAR~\cite{qiu2019deeplidar}   & 758.38 & 226.50 & 2.56 & 1.15  \\
    		CSPN++~\cite{cheng2020cspn++}  & 743.69  & 209.28 & {2.07}  & \textcolor{red}{0.90}  \\
          	GuideNet~\cite{tang2019learning}& 736.24  & 218.83 & 2.25  & 0.99 \\
                FCFRNet~\cite{liu2021fcfr} & 735.81  & 217.15 & 2.20  & 0.98 \\
                ACMNet~\cite{zhao2021adaptive}  & 732.99  & {206.80} & 2.08  & 0.90 \\
    		PENet~\cite{HuWLNFG21}   & 730.08  & 210.55 & 2.17  & 0.94 \\
                GuideFormer~\cite{rho2022guideformer}  & 721.48  & 207.76 & 2.14  & 0.97 \\
            \hline
          	\myrowcolour Our \emph{DGDF-A}    & \textcolor{blue}{\textbf{708.30}}  & \textcolor{blue}{\textbf{205.01}} & \textcolor{red}{\textbf{2.04}}  & \textcolor{blue}{\textbf{0.91}} \\
                \myrowcolour Our \emph{DGDF-B}    & \textcolor{red}{\textbf{707.93}}  & \textcolor{red}{\textbf{205.11}} & \textcolor{blue}{\textbf{2.05}}  & \textcolor{blue}{\textbf{0.91}} \\
    		\hline
\end{tabular}
\vspace{-3mm}
\end{table}

\subsection{Experiments on KITTI dataset}

To verify the performance of the proposed methods, we compare our methods with other state-of-the-art (SOTA) depth completion methods on the KITTI benchmark~\cite{Geiger2012CVPR} qualitatively and quantitatively.

\noindent\textbf{Quantitative comparisons.}
Table~\ref{tab:Quantitative} shows the quantitative evaluation of our methods and other SOTA methods on the KITTI leaderboard that ranks all methods according to the RMSE metric.
The experimental results demonstrate that our methods achieve strong results.
Our methods \emph{DGDF-A} and \emph{DGDF-B} outperform other SOTA methods under the primary RMSE metric.
\emph{DGDF-B} and \emph{DGDF-A} rank 1st and 2nd at the time of paper submission, respectively.
In addition, \emph{DGDF-A} and \emph{DGDF-B} also achieve comparable performance under other metrics.
Specifically, existing depth completion methods, such as S2D~\cite{MaCK19}, Uncertainty~\cite{van2019sparse}, DeepLiDAR~\cite{qiu2019deeplidar} and PENet~\cite{HuWLNFG21}, employ standard or modified encoder-decoder structures to regress the sparse depth maps and corresponding RGB images to the dense depth maps, where the dense RGB images and sparse LiDAR depth are fused by element-wise addition or channel-wise concatenation.
These simple feature fusion methods are not able to fully exploit the potential of RGB images as guidance.
Although GuideNet~\cite{tang2019learning} addresses this issue by guided dynamic filters and achieves good results, the feature fusion method suffers from heavy model parameters and hardware costs.
Our methods efficiently integrate the RGB images and sparse depth maps by the proposed decomposed guided dynamic filters, they achieve satisfactory results.
Meanwhile, since the sparse depth maps are obtained by projecting LiDAR point clouds to the image plane, the position displacement between LiDAR and the camera will inevitably cause that some foreground and background points are overlapped in the depth map.
GAENet~\cite{du2022depth} and ACMNet~\cite{zhao2021adaptive} address this issue by introducing the geometry information.
Although our methods are agnostic to the geometry, they still achieve good results, which explicitly utilize the RGB information to guide depth features in the overlapped areas. 
In addition, we also compare our methods with a series of SPN-based methods that use the spatial propagation network (SPN) to refine the predicted depth map.
Our methods do not employ additional refine modules, but they still perform better than some such methods, such as CSPN~\cite{cheng2019cspn} and CSPN++~\cite{cheng2020cspn++}.

\begin{table}[]
\renewcommand\arraystretch{1.2}
\centering
\caption{Quantitative comparison with state-of-the-art depth completion methods on the NYUv2 dataset. The best and second-best RMSE are highlighted in \textcolor{red}{red} and \textcolor{blue}{blue} colors, respectively.}
\label{tab:NYUv2}
\begin{tabular}{lcccccc@{}}
\hline 
\myrowcolour Methods   & \begin{tabular}[c]{@{}c@{}}RMSE \\ {[}m{]}\end{tabular} &  REL  & $\delta_{1.25}$ & $\delta_{{1.25}^{2}}$ &  $\delta_{{1.25}^{3}}$ \\ \hline 
TGV~\cite{ferstl2013image}  & 0.635 & 0.123 & 81.9 & 93.0 & 96.8 \\
Bilateral~\cite{barron2016fast}  & 0.479 & 0.084 & 92.4 & 97.6 & 98.9 \\
S2D~\cite{MaCK19}  & 0.230 & 0.044 & 97.1 & 99.4 & 99.8  \\
CSPN~\cite{cheng2019cspn}  & 0.117 & 0.016 & 99.2 & 99.9 & 100.0 \\
DeepLiDAR~\cite{qiu2019deeplidar}  & 0.115 & 0.022 & 99.3 & 99.9 & 100.0 \\
DepthNormal~\cite{xu2019depth}  & 0.112 & 0.018 & 99.5 & 99.9 & 100.0 \\
FCFRNet~\cite{liu2021fcfr}  & 0.106 & {0.015} & 99.5 & 99.9 & 100.0 \\
ACMNet~\cite{zhao2021adaptive}   & 0.105 & {0.015} & 99.4 & 99.9 & 100.0 \\
PRNet~\cite{lee2021depth}     & 0.104 & {0.014} & 99.4 & 99.9 & 100.0 \\
GuideNet~\cite{tang2019learning} & 0.101 & {0.015} & 99.5 & 99.9 & 100.0 \\
\hline
\rowcolor{lightgray}
\myrowcolour Our \emph{DGDF-A}    & \textcolor{blue}{\textbf{0.099}}  &{\textbf{0.014}} & \textbf{99.5} & \textbf{99.9} & \textbf{100.0} \\
\rowcolor{lightgray}
\myrowcolour Our \emph{DGDF-B}    & \textcolor{red}{\textbf{0.098}}  & {\textbf{0.014}} & \textbf{99.5} & \textbf{99.9} & \textbf{100.0} \\
\hline 
\end{tabular}
\end{table}

\noindent\textbf{Qualitative comparisons.}
Fig.~\ref{fig:Qualitative} compares the dense depth maps predicted by our \emph{DGDF-A} and \emph{DGDF-B} and other state-of-the-art methods, such as GuideNet~\cite{tang2019learning}, CSPN~\cite{cheng2019cspn}, and ACMNet~\cite{zhao2021adaptive}.
As shown in the enlarged areas of the first column, our \emph{DGDF-A} and \emph{DGDF-B} recover better details.
Specifically, the dense depth maps estimated by GuideNet~\cite{tang2019learning} and ACMNet~\cite{zhao2021adaptive} produce ripples similar to water waves, and the depth maps predicted by CSPN~\cite{cheng2019cspn} are shifted at the boundary of the car.
While the results of our \emph{DGDF-A} and \emph{DGDF-B} are more accurate at the object boundaries.
In addition, we can observe the same results in the enlarged areas of the second column, the predicted depth maps of our \emph{DGDF-A} and \emph{DGDF-B} present the most accurate contour at the boundary of the tree.

\begin{figure*}[!t]
	\centering
	\includegraphics[width=0.92\textwidth]{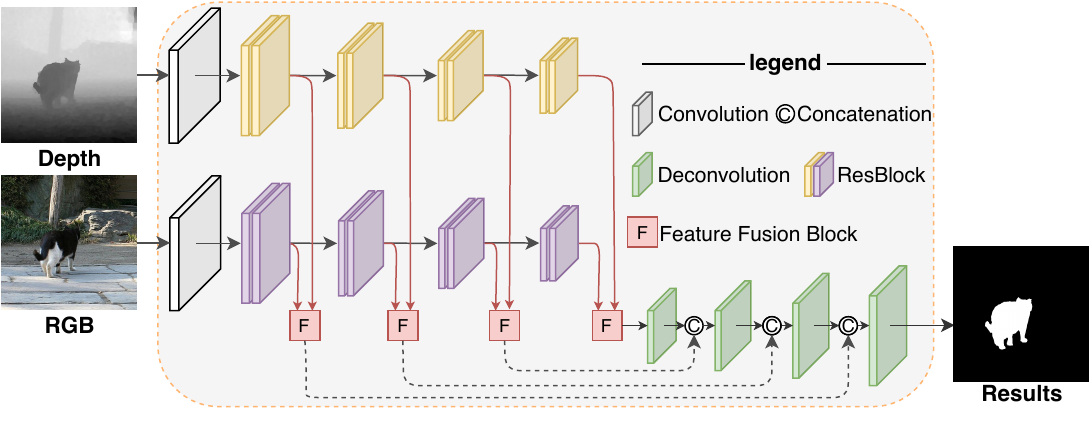}
        \vspace{-1mm}
	\caption{The RGB-D salient object detection network architecture employed by our proposed method, which extracts multi-scale RGB features and depth features by two separate networks and fuses them through our proposed methods.}
	\label{fig:network_sod}
	\vspace{-2mm}
\end{figure*}

\begin{figure*}[htbp]
\centering
\includegraphics[width=1.0\textwidth]{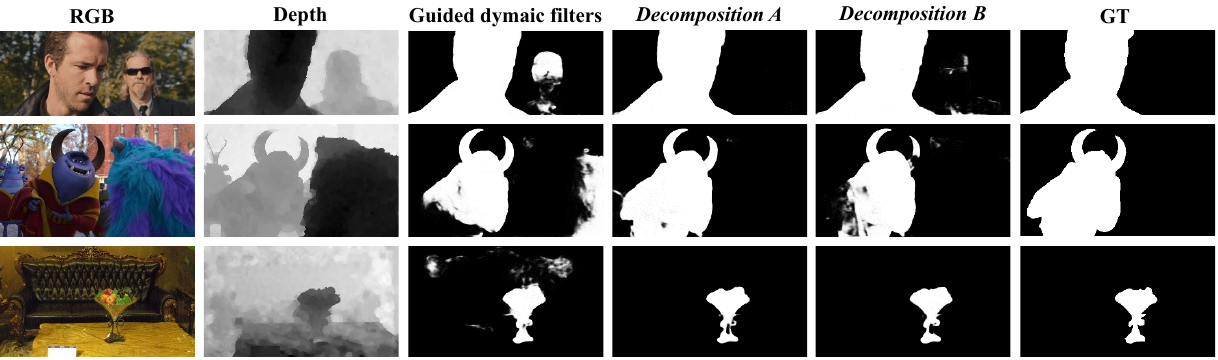}
\caption{{Typical examples in the RGB-D salient object detection task. From left to right are RGB images, depth maps, the results predicted by our RGB-D SOD method using different feature fusion methods, including guided dynamic filters~\cite{tang2019learning} and our decomposed guided dynamic filters $\mathcal{A}$ and $\mathcal{B}$, and the ground truth.}}
\label{fig:sod_results}
\vspace{-2mm}
\end{figure*}

\subsection{Experiments on the NYUv2 dataset}
To verify the generalization ability of our methods in the indoor scenes, we conduct extensive experiments on the NYUv2 dataset.
We generate the sparse input depth maps by randomly sampling from the dense ground truth.
Following the existing methods, our models are trained and tested under the setting of 500 sparse LiDAR samples.
In addition, for a fair comparison, we pad the input images to $320\times 256$ but evaluate only the valid region of size $304\times 228$ as GuideNet~\cite{tang2019learning}.
The quantitative results in \tabref{tab:NYUv2} demonstrate that our methods \emph{DGDF-A} and \emph{DGDF-B} achieve a consistent improvement over GuideNet on all metrics and show comparable performance with other state-of-the-art methods.
For the NYUv2 dataset, although only 0.6\% of the pixels in the sparse depth map have values, which are more sparse than the sparse depth map of the KITTI dataset (4\% of pixels have depth values), our methods still predict the dense depth maps well by fully exploiting the ability of RGB images as guidance.
The qualitative results in Fig.~\ref{fig:nyuv2} demonstrate that, compared with the GuideNet, our methods preserve tiny structures and depth boundaries better.

\begin{table*}[!t]
\renewcommand\arraystretch{1.1}
  \centering
  \renewcommand{\arraystretch}{1.0}
  \renewcommand{\tabcolsep}{1mm}
  \caption{{Performance comparison of our RGB-D SOD method using different feature fusion methods.
  The best and second-best results are highlighted in \textcolor{red}{red} and \textcolor{blue}{blue} colors, respectively.
  }}
  \resizebox{\textwidth}{!}{%
  \begin{tabular}{c|cccc|cccc|cccc|cccc|cccc}
  \hline
  \myrowcolour &\multicolumn{4}{c|}{NJU2K \cite{NJU2000}}&\multicolumn{4}{c|}{SSB \cite{niu2012leveraging}}&\multicolumn{4}{c|}{DES \cite{cheng2014depth}}&\multicolumn{4}{c|}{NLPR \cite{peng2014rgbd}}&\multicolumn{4}{c}{LFSD \cite{li2014saliency}} \\
    \myrowcolour Method & $S_{\alpha}\uparrow$&$F_{\beta}\uparrow$&$E_{\xi}\uparrow$&$\mathcal{M}\downarrow$& $S_{\alpha}\uparrow$&$F_{\beta}\uparrow$&$E_{\xi}\uparrow$&$\mathcal{M}\downarrow$& $S_{\alpha}\uparrow$&$F_{\beta}\uparrow$&$E_{\xi}\uparrow$&$\mathcal{M}\downarrow$& $S_{\alpha}\uparrow$&$F_{\beta}\uparrow$&$E_{\xi}\uparrow$&$\mathcal{M}\downarrow$& $S_{\alpha}\uparrow$&$F_{\beta}\uparrow$&$E_{\xi}\uparrow$&$\mathcal{M}\downarrow$  \\ \hline

No Depth~ &.909 &.900 &.940 &.036 &.897 &.880 &.936 &.039 &.938 &.927 &.974 &.018 &.921 &.899 &.957 &.023 &.814 &.813 &.853 &.090   \\ 
Add &.912 &.904 &.941 &.035 &.901 &.883 &\textcolor{blue}{.938} &\textcolor{blue}{.038} &\textcolor{blue}{.939} &\textcolor{blue}{.929} &\textcolor{red}{.977} &\textcolor{red}{.015} &.923 &.902 &\textcolor{blue}{.959} &.023 &.829 &.823 &\textcolor{blue}{.866} &.084   \\
Concatenation &.913 &.906 &.944 &.034 &.900 &\textcolor{blue}{.884} &.937 &.039 &.935 &\textcolor{blue}{.929} &.971 &\textcolor{blue}{.016} &.921 &.901 &.957 &.023 &\textcolor{blue}{.830} &\textcolor{blue}{.832} &\textcolor{blue}{.866} &\textcolor{blue}{.083}   \\
Guide dynamic filters~\cite{tang2019learning} &\textcolor{blue}{.919} &\textcolor{blue}{.912} &\textcolor{blue}{.948} &.033 &\textcolor{red}{.903} &.883 &\textcolor{blue}{.938} &\textcolor{blue}{.038} &.938 &\textcolor{blue}{.929} &.973 &.017 &\textcolor{blue}{.924} &\textcolor{blue}{.904} &.957 &.023 &.822 &.817 &.857 &.088   \\
    \hline
\myrowcolour Decomposition $\mathcal{A}$ &\textcolor{red}{.921} &\textcolor{red}{.914} &\textcolor{red}{.949} &\textcolor{red}{.031} &\textcolor{red}{.903} &\textcolor{red}{.886} &.936 &\textcolor{red}{.037} &.936 &.926 &.968 &.018 &\textcolor{red}{.926} &\textcolor{red}{.908} &\textcolor{red}{.961} &\textcolor{red}{.021} &\textcolor{red}{.842} &\textcolor{red}{.834} &\textcolor{red}{.872} &\textcolor{red}{.078}   \\
\myrowcolour Decomposition $\mathcal{B}$  &.917 &.909 &.947 &\textcolor{blue}{.032} &\textcolor{blue}{.902} &\textcolor{blue}{.884} &\textcolor{red}{.939} &\textcolor{blue}{.038} &\textcolor{red}{.942} &\textcolor{red}{.930} &\textcolor{blue}{.975} &\textcolor{red}{.015} &.921 &.901 &.958 &\textcolor{blue}{.022} &.823 &.818 &.856 &.086   \\
    \hline
    \end{tabular}
    }
    \label{tab:benchmark_rgbd_sod}
    \vspace{-2mm}
\end{table*}

\subsection{Extension to RGB-D salient object detection}
Our proposed decomposed guided dynamic filters are general, which can not only help the RGB-guided depth completion methods achieve SOTA performance, but also boost other multi-modal input tasks.
In this subsection, we extend our proposed decomposed guided dynamic filters to the RGB-D salient object detection (SOD) task~\cite{Wang2022sod, Liu2022sod} to fuse the features from RGB images and depth maps.

The network architecture of our RGB-D salient object detection method is shown in Fig.~\ref{fig:network_sod}. We adopt the ResNet50~\cite{he_ResNet_CVPR_2016} as our backbone, which takes RGB images and depth maps as input and produces two lists of feature maps $f_{\theta_{1}}(x^{R})=\{t_l^{R}\}_{l=1}^4$, $f_{\theta_{2}}(x^{D})=\{t_l^{D}\}_{l=1}^4$, representing different levels of the features from the RGB image $R$ and depth map $D$. 
Then, we feed each backbone feature $t_l^{R}$ and $t_l^{D}$ to a simple convolution layer and obtain the new backbone feature $\{s_l^{R}\}_{l=1}^4$ and $\{s_l^{D}\}_{l=1}^4$ of same channel size $C=64$. 
Furthermore, we perform feature fusion at each level and obtain fused features $\{s_l^{F}\}_{l=1}^4$. 
Finally, we employ a UNet~\cite{ronneberger_Unet_2015} decoder to decode the fused feature and get the saliency prediction. We follow the implementation and evaluation details as~\cite{mao_TransformerSOD_arxiv_2021}. We compare the performance between the method not using depth maps and the methods using different feature fusion methods, including the element-wise addition (Add), the channel-wise concatenation (Concatenation), guided dynamic filters~\cite{tang2019learning} and our proposed decomposed guided dynamic filters $\mathcal{A}$ and $\mathcal{B}$.
Quantitative results are reported in Table~\ref{tab:benchmark_rgbd_sod}.
We observe that the methods using depth maps present better results, as the depth maps provide additional useful information.
Meanwhile, the methods using our proposed decomposed guided dynamic filters $\mathcal{A}$ and $\mathcal{B}$ achieve consistent improvement compared with the methods using other feature fusion methods. 
In addition, we show some qualitative results in Fig.~\ref{fig:sod_results}, which demonstrate that our methods improve the quality of the salient object detection results.
In summary, both the qualitative and quantitative experimental results demonstrate that our proposed methods are also effective in the RGB-D salient object detection task.

\section{Conclusion}\label{conclusion}
In this paper, we have proposed the decomposed guided dynamic filters.
Instead of directly generating a complete depth-wise convolutional filter at each spatial location, our key insight is to reconstruct the spatially-variant and content-adaptive filters by multiplying the spatially-shared component with content-adaptive adaptors. 
Along this pipeline, we proposed two decomposition schemes $\mathcal{A}$ and $\mathcal{B}$.
Our proposed methods are effective and efficient, which significantly reduce model parameters, computational cost, and memory footprints, while helping the depth completion methods achieve state-of-the-art performance.
In addition, the proposed methods could be used as plug-and-play feature fusion blocks to boost other multi-modal fusion tasks. Extended experiments on the RGB-D salient object detection task demonstrate that our methods can effectively integrate the multi-modal information to improve the performance of the method.

\bibliographystyle{unsrt}
\bibliography{ddfnetbib_v2}

\end{document}